\documentclass[sigconf]{acmart}
\AtBeginDocument{%
  }

\setcopyright{rightsretained}
\copyrightyear{2025}
\acmYear{2025}
\acmDOI{XXXXXXX.XXXXXXX}
\acmConference[AAE '25]{Workshop on Agentic AI for Enterprise at KDD '25}{August 4,
  2025}{Toronto, Canada}
\acmBooktitle{AAE '25: Workshop on Agentic AI for Enterprise at KDD '25,
 August 4, 2025, Toronto, Canada}
\acmISBN{XXX-X-XXXX-XXXX-X/2025/08}

\usepackage{listings}
\usepackage{array}
\usepackage{enumitem}
\usepackage{float}
\usepackage{algorithm}
\usepackage{algpseudocode}
\usepackage{amsmath}
\usepackage{booktabs}

\begin{document}

\title{Text-to-SQL for Enterprise Data Analytics}


\author{Albert Chen}
\authornote{Correspondence to abchen@linkedin.com.}

\author{Manas Bundele}

\author{Gaurav Ahlawat}

\author{Patrick Stetz}
\affiliation{%
  \institution{LinkedIn}
  \city{Sunnyvale}
  \state{California}
  \country{USA}
}

\author{Zhitao Wang}

\author{Qiang Fei}

\author{Donghoon Jung}

\author{Audrey Chu}
\affiliation{%
  \institution{LinkedIn}
  \city{Sunnyvale}
  \state{California}
  \country{USA}
}
\authornote{Work done while at LinkedIn.}

\author{Bharadwaj Jayaraman}

\author{Ayushi Panth}

\author{Yatin Arora}

\author{Sourav Jain}

\author{Renjith Varma}
\affiliation{%
  \institution{LinkedIn}
  \city{Sunnyvale}
  \state{California}
  \country{USA}
}

\author{Alexey Ilin}

\author{Iuliia Melnychuk}

\author{Chelsea Chueh}

\author{Joyan Sil}

\author{Xiaofeng Wang}
\affiliation{%
  \institution{LinkedIn}
  \city{Sunnyvale}
  \state{California}
  \country{USA}
}

\renewcommand{\shortauthors}{Chen et al.}

\begin{abstract}
    The introduction of large language models has brought rapid progress on Text-to-SQL benchmarks, but it is not yet easy to build a working enterprise solution. In this paper, we present insights from building an internal chatbot that enables LinkedIn's product managers, engineers, and operations teams to self-serve data insights from a large, dynamic data lake. Our approach features three components. First, we construct a knowledge graph that captures up-to-date semantics by indexing database metadata, historical query logs, wikis, and code. We apply clustering to identify relevant tables for each team or product area. Second, we build a Text-to-SQL agent that retrieves and ranks context from the knowledge graph, writes a query, and automatically corrects hallucinations and syntax errors. Third, we build an interactive chatbot that supports various user intents, from data discovery to query writing to debugging, and displays responses in rich UI elements to encourage follow-up chats. Our chatbot has over 300 weekly users. Expert review shows that 53\% of its responses are correct or close to correct on an internal benchmark set. Through ablation studies, we identify the most important knowledge graph and modeling components, offering a practical path for developing enterprise Text-to-SQL solutions.
\end{abstract}



\settopmatter{printfolios=true}
\maketitle

\section{Introduction}
Deploying an industry-scale Text-to-SQL chatbot requires more than just a model proficient at writing SQL. Key requirements include understanding domain-specific semantics, generalizing to a large and evolving data lake, and delivering end-user utility. In this paper we highlight our approach to make data insights accessible to any employee at LinkedIn, with a focus on SQL novices who have familiarity with business intelligence charts but little-to-no familiarity with the raw data. An important requirement is that our chatbot must work for the whole company, not limited to a particular business vertical. We focus on Trino SQL \cite{trino}.

To illustrate the challenges, our data lake contains millions of tables, with some popular tables having over 100 columns. Tables are regularly deprecated and many tables contain overlapping information. Responses must be personalized to each user: a request for the latest click-through rate has a different meaning for an employee working in notifications compared to one working in search. The chatbot must understand business jargon and acronyms that are unique to our company. And our users expect the chatbot to assist with auxiliary tasks in the query-authoring process (e.g. find tables, explain queries).

There has been a rapid improvement in public Text-to-SQL benchmarks such as Spider \cite{yu-etal-2018-spider} and BIRD \cite{li2023llmservedatabaseinterface}. Their primary evaluation metric is execution accuracy, defined as the percent of queries that produce output identical to the ground truth SQL. The Spider leaderboard shows an improvement in execution accuracy from 54\% in May 2020 \cite{zhong-etal-2020-grounded} to over 90\% in November 2023 \cite{spider_leaderboard}. BIRD was introduced in 2023 to provide more challenging Text-to-SQL tasks. Its leaderboard also shows rapid improvement in execution accuracy from 40\% in March 2023 \cite{li2023llmservedatabaseinterface} to 76\% by April 2025 \cite{bird_leaderboard}. This progress is largely due to recent advances in large language models (LLMs). However, such rapid progress does not immediately translate to enterprise settings. \citet{floratou_nl2sql_2024} notes challenges in handling large enterprise schemas and semantic ambiguity. The Spider 2.0 benchmark, introduced in November 2024, is a more realistic benchmark, with ground truth queries over 100 lines long on tables with over 1000 columns \cite{lei2024spider20evaluatinglanguage}. In April 2025, the best model's execution accuracy was only 31\% \cite{spider2_leaderboard}. Similarly, Uber built a internal Text-to-SQL application and report ~50\% overlap with ground truth tables on their own evaluation set \cite{query_gpt_uber}.

State-of-the-art Text-to-SQL approaches involve context selection followed by query generation and then query correction. Context selection primarily involves schema linking, which is the task of mapping terms in the user query to tables, columns, and values in the database. This is done through some combination of entity extraction, embedding-based retrieval (EBR), LLMs, and locality-sensitive hashing \cite{pourreza_chase-sql_2024, talaei_chess_2024}. It also includes retrieval of relevant example queries for in-context learning. Query generation then relies on passing the appropriate context to an LLM to generate a query. Query generation for complex tasks can be improved by breaking down the query into sub-tasks to be solved sequentially \cite{pourreza_din-sql_2023}. Another approach is to generate multiple queries and select the best one via self-consistency \cite{gao_text--sql_2023, talaei_chess_2024} or a selection model \cite{pourreza_chase-sql_2024}. Query correction then fixes errors based on query execution output or self-reflection \cite{maamari_death_2024, pourreza_din-sql_2023, talaei_chess_2024}.

Our approach builds on these methods, focusing on the three key contributions to address enterprise challenges:
\begin{enumerate}[leftmargin=*]
    \item To understand data semantics, we construct a knowledge graph from table schemas, documentation, code repos, historical query logs, company jargon, and crowdsourced domain knowledge. To organize the data lake, we cluster tables and associate table clusters with users and product areas. The knowledge graph is regularly refreshed to stay up-to-date.
    \item To write queries, the Query Writer Agent uses multi-stage retrieval and ranking that identifies the most useful tables, columns, examples, and other context. It includes a Researcher LLM Agent to retrieve additional information to fix table / column hallucination.
    \item To deliver end-user utility, we design an interactive chat UI that helps users understand the query and reply to the bot. Interactivity allows users to clarify ambiguities and improve bot responses. However, speed is critical for maintaining a seamless chat experience. We take a multi-agent approach to ensure efficient follow-ups and handle diverse questions beyond Text-to-SQL.
\end{enumerate}

To our knowledge, this paper is the first detailed presentation of an enterprise Text-to-SQL solution.
\section{Methodology}

\subsection{Knowledge Graph}
The Text-to-SQL task involves identifying which tables to query and correctly querying them to get the answer. Our knowledge graph organizes information from both databases and users to aid in this task. User-generated information includes table/column documentation, historical query logs, code, wiki pages, and domain knowledge collected in our chatbot UI. The domain knowledge includes product area background, data explanations, and user preferences. These sources are unique to the enterprise setting and contain valuable semantic information. The knowledge graph is used to retrieve context for query generation and validation.

\begin{figure}[h]
    \centering
    \includegraphics[width=0.9\linewidth]{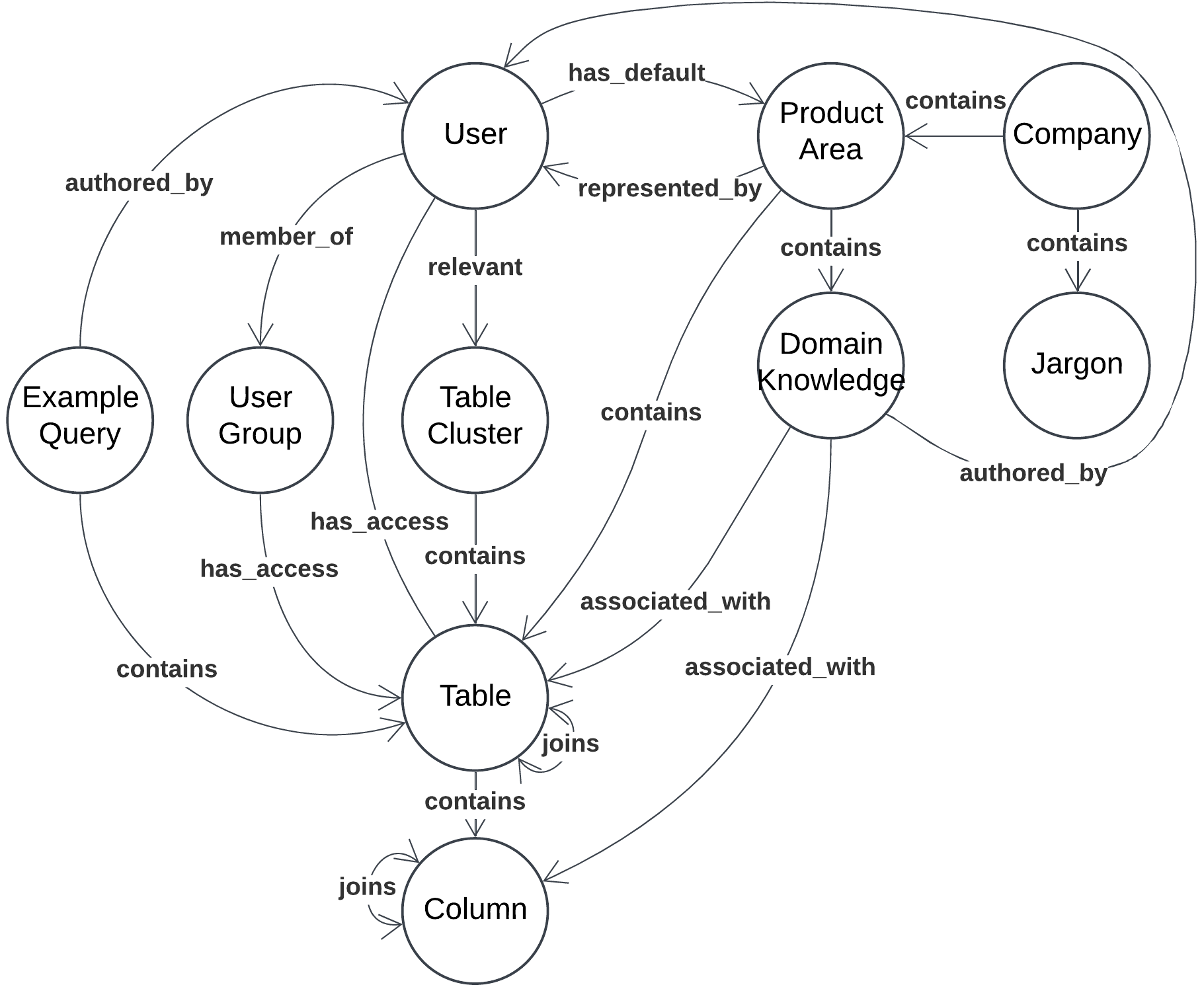}
    \caption{Knowledge graph for Text-to-SQL.}
    \label{fig:knowledge_graph}
    \Description[Nodes include table, table cluster, user, product area, company]{Nodes include table, table cluster, user, product area, company, domain knowledge. Many of the nodes have an edge to the table node. For example, domain knowledge is associated with a table.}
\end{figure}

The nodes and edges of our knowledge graph are shown in Figure~\ref{fig:knowledge_graph}. Tables are the central organizing entity. Given a table, we can retrieve information about its semantics: columns, common joins, example queries, and domain knowledge records. We can also retrieve the users or groups who have permission to read from the table. The other organizing entity is the product area. The company contains multiple product areas, each with respective domain knowledge records. Users and product areas are associated with tables through a clustering approach (see Sec~\ref{sec:clustering}).

Table~\ref{tab:knowledge_graph} shows the attributes for tables and columns. These attributes provide much richer semantic information than the schema alone; we use them to provide context to users and LLMs. For example, certified tables are high quality tables recommended for data consumers to use. Column type indicates whether a column is a metric, dimension, or attribute, as annotated by the dataset owner. Popular tables and columns have human-authored descriptions, supplemented by LLM-authored descriptions based on schemas and wikis using Glean's Chat API \cite{glean_chat_api}.

\begin{table}[h]
\centering
\small 
\begin{tabular}{| p{1cm} | p{6cm} |}
\hline
Node & Attributes \\
\hline
Table & Database Name, Table Name, Human Description, AI Description, Usage Popularity, Table Cluster, Tags, Certification Status, Deprecation Status \\
\hline
Column & Database Name, Table Name, Column Name, Human Description, AI Description, Usage Popularity, Top Values, Data Type, Column Type, Is Partition Key \\
\hline

\end{tabular}
\caption{Attributes of the table and column nodes.}
\label{tab:knowledge_graph}
\end{table}

\subsubsection{User-Dataset Clustering}
\label{sec:clustering}
We have millions of tables but not all are relevant to every product area or user. We take a clustering approach based on user-table access counts from three months of query history. Clustering pseudocode is provided in Appendix~\ref{ref:clustering_pseudocode}. We first filter this matrix to the tables with a sufficient number of total and unique user accesses, reducing noise and removing tables that are rarely accessed (e.g. intermediate tables produced by data pipelines). Such tables would be difficult to cluster meaningfully.

Next, we perform dimensionality reduction across the user dimension using Independent Component Analysis (ICA) with $N_{\text{comp}}$ components \cite{HYVARINEN2000411,pedregosa2011scikit}. This captures variations in tables access patterns. ICA produces a score for each (table, component) pair indicating the strength of their association. We use this to produce a soft clustering of tables: for each component, we take the top $T_{\text{c}}$ tables with the strongest scores. We also assign each table to its component with the highest score. The result is $N_{\text{comp}}$ clusters with at least $T_{\text{c}}$ tables each where $N_{\text{comp}}=200$ and $T_{\text{c}}=20$ based on manual review of the table clusters generated under different settings. Tables per cluster $T_{\text{c}}$ should be high enough to include closely related tables in a single cluster. The number of clusters $N_{\text{comp}}$ should be high enough to cover different table access patterns across the business.

After clustering, we find the top clusters for each user by summing their access counts in each component, and assigning the top components as the personalized components for that user. Next, we use team email groups to identify representative employees for each product area. We identify the top clusters for a product area as those associated with the most team members, breaking ties by cluster access count. Tables can also be directly added to a product area. Our soft clustering approach allows a table to belong to multiple clusters, and a cluster to belong to multiple product areas, allowing them to be shared across products and adapting to changes like table deprecation or team switching by users. By associating users and product areas with table clusters, we include similar tables that a user or team has not directly accessed before.

The end-to-end clustering process only takes 15 minutes to run (p90). Clustering defines and organizes the default set of tables used by the chatbot to write queries. To support the long tail, we dynamically retrieve additional tables within a chat session if requested.

\subsubsection{Knowledge Graph Indexing}
The knowledge graph is refreshed regularly to keep up-to-date with the latest semantics and datasets. We store it across multiple indexes for efficient update and retrieval.

\textbf{Table/column indexes}. Contains tables and columns sourced from an internal instance of DataHub \cite{datahub}, along with their attributes. We embed the table name, descriptions and tags (human-generated labels). Supports embedding-based retrieval (EBR) for tables and retrieval by table name/column name/product area; refreshed weekly.

\textbf{Table/column usage indexes}. Contains table and column information derived from historical query logs, such as popularity and common joins. We parse the Trino \texttt{EXPLAIN} plan (i.e. the logical query plan in json format) from successful queries to extract fully qualified names of tables, columns and any join conditions \cite{trino_explain}. We aggregate to identify usage popularity and common joins in a recent period. Supports retrieval by table/column name; refreshed weekly.

\textbf{Table cluster index}. Contains a list of relevant datasets for each user. Supports retrieval by user; refreshed weekly.

\textbf{Example query index}. Contains human-authored example queries sourced from code repos and wikis. Examples from wikis have human-authored descriptions which are embedded for retrieval. Queries from code repos are filtered to keep high-quality queries based on the creation date, execution count, filename, and whether the query is certified by the user. We use \texttt{gpt-35-turbo} to generate descriptions and embed the description and the query for retrieval \cite{open_ai_gpt_35_turbo}. Supports embedding-based retrieval and filtering by table or author; refreshed as needed.

\textbf{Domain knowledge index}. Contains user-supplied knowledge text with dimensions for product areas, tables, columns. These are collected in the chatbot UI and can include product background information, data explanations, and user preferences. Supports retrieval by product area, table, column, or user; refreshed instantly. Jargon is sourced from company wiki pages and stored as a map from jargon to explanation.

\subsection{Query Writer Agent}

\begin{figure}[h]
    \vspace{-0.1in}
    \centering
    \includegraphics[width=0.9\linewidth]{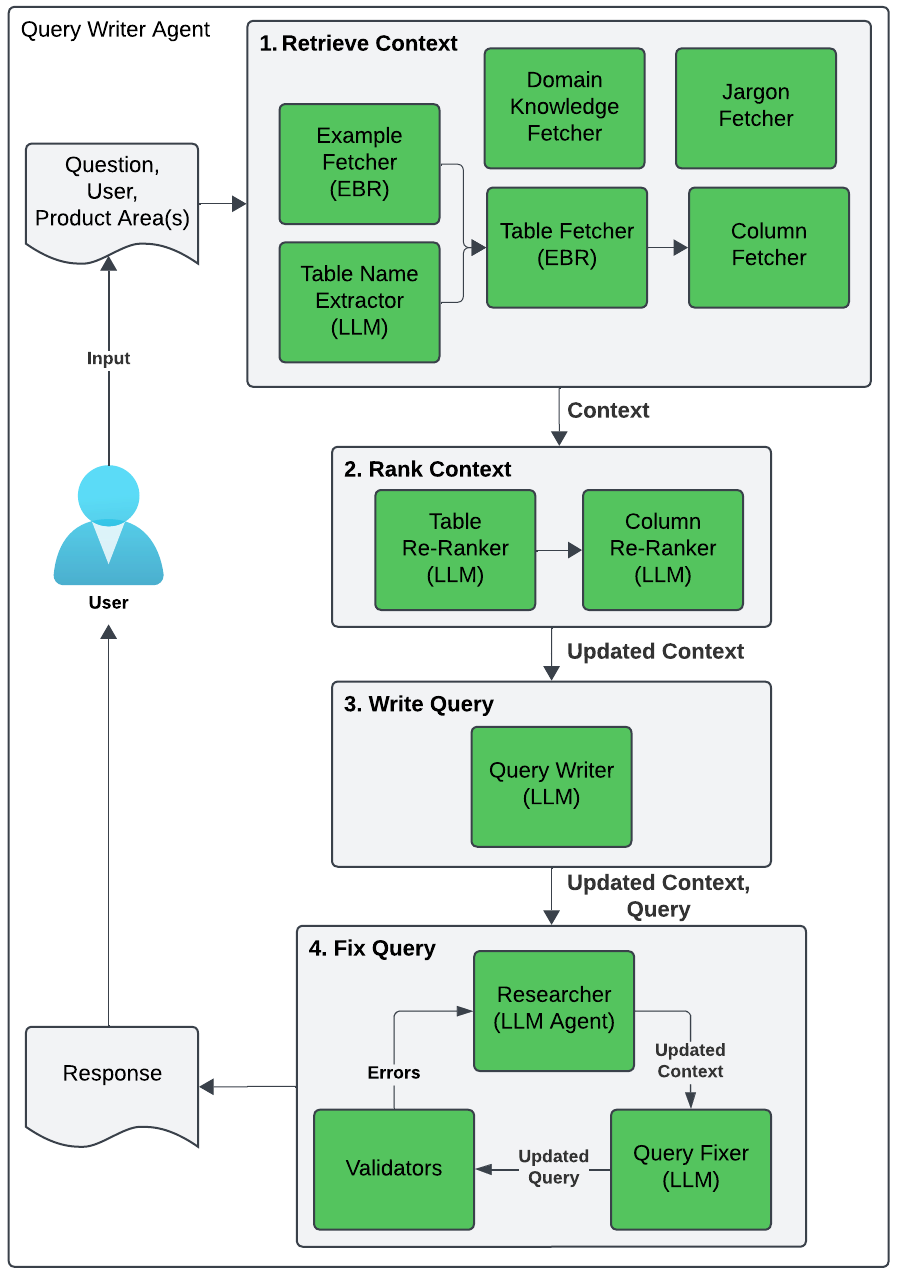}
    \caption{Query Writer Agent architecture. The Agent retrieves and ranks context from the knowledge graph, then writes and fixes the query. The fixer is equipped with a Researcher LLM Agent that can retrieve additional information from the knowledge graph to fix hallucination errors.}
    \label{fig:query_writer_agent}
    \Description[The user's question is sent to the query writer agent for a response.]{The query writer agent has four stages. Retrieve context passes context to Rank context, which passes context to Write query, which passes updated context and query to Fix query. Within each step are boxes showing the work involved. For example, Retrieve Context starts with example fetcher (EBR) and table name extractor (LLM), followed by Table Fetcher (EBR) and Column Fetcher. Domain knowledge and Jargon are also retrieved.}
    \vspace{-0.1in}
\end{figure}

In this section, we present the end-to-end query writing approach used by the Query Writer Agent, as illustrated in Figure~\ref{fig:query_writer_agent}. The Query Writer Agent has four steps: retrieving context, ranking context, writing a query, and fixing the query. These steps are used in state-of-the-art approaches, although implementation details are different and not all steps are included in every approach \cite{talaei_chess_2024,pourreza_chase-sql_2024,pourreza_din-sql_2023,gao_text--sql_2023}. We implemented this agent using \texttt{langchain} \cite{langchain}.

The inputs are the question, user name, and a list of product areas. The agent retrieves initial context from the knowledge graph (tables, example queries, domain knowledge, etc.) and refines the context through subsequent ranking, writing, and fixing steps. For example, the Researcher Agent can retrieve additional tables to address hallucination issues. Each LLM's prompt is generated dynamically using the current context.

\subsubsection{Retrieve Context}
Incorrect or irrelevant table schemas in the prompt \cite{floratou_nl2sql_2024} hurt the quality of the answer. The retrieval stage uses lightweight techniques to retrieve a subset of context from the knowledge graph, with the objective of high recall. First, a set of candidate tables is initialized using the table clusters for the user and selected product areas. Examples are retrieved via semantic similarity to the user's question, discarding those that do not use a candidate table. Next, it retrieves top $K_{\text{ret}}$ tables to pass to ranking. The tables are pulled from three sources: (1) Embedding-based retrieval (EBR) using the user's question to query the table index, (2) tables in the retrieved examples, (3) tables in the user's question. Results from (1) and (2) are limited to the candidate tables, whereas (3) provides flexibility to use any user-mentioned table. Finally, it fetches all the columns for the top $K_{\text{ret}}$ tables, domain knowledge records for selected product areas, and jargon explanations via string matching. We set $K_{\text{ret}}=20$ to maximize recall while avoiding too many LLM calls for the following ranking step.

\subsubsection{Rank Context}
The context ranking step selects the top $K_{\text{rnk}}$ tables and a subset of relevant columns to use for query writing. While the previous step uses lightweight EBR to retrieve context across indexes independently, context ranking uses a LLM to reason across all retrieved information. First, the table ranker LLM scores the relevance of each table to the question (on a scale from 1-10, with explanations) to identify the top $K_{\text{rnk}}$ tables using table names, descriptions, example queries, commonly joined or co-queried tables, domain knowledge, and explanations of internal jargon. Using table schemas lowered recall and increased latency, so we omit them at this step. Next, the column ranker identifies a subset of columns from the top $K_{\text{rnk}}$ tables to pass to the query writer, separated into two tiers: relevant and potentially relevant. Including the second tier improves recall. The column ranker LLM is given similar context as the table ranker, limited to the top $K_{\text{rnk}}$ tables. Full table schemas are provided with extended column descriptions generated from the attributes in Table~\ref{tab:knowledge_graph}, ordered by usage popularity. Each table is represented by a CREATE TABLE statement, similar to the code representation prompt in \citet{gao_text--sql_2023}, but with extended descriptions. Choosing $K_{\text{rnk}}$ balances precision and recall: higher values improves context ranker's recall, but also increases token usage, slowing down the query writer and leading to errors. $K_{\text{rnk}}=7$ provides good accuracy in our case.

\subsubsection{Write and Fix Query}
The input prompt for the query writer LLM contains tables and columns selected after ranking, table relevance scores and explanations, examples, domain knowledge, and jargon. It returns assumptions, query, explanation, tables, and columns. After query writing, the query validation loop runs at most twice to address compilation and hallucination issues. For hallucination, we focus on invalid tables and columns, which are easy to detect and affect the query structurally, and leave invalid column values to future work. We detect errors by running Trino \texttt{EXPLAIN} in \texttt{VALIDATE} mode, which validates syntactical and semantic correctness (e.g. syntax, function usage, table name, column name). Since Trino \texttt{EXPLAIN} only returns one error at a time, we run another hallucination validator to detect all table and column hallucination issues so the errors can be fixed together. It checks extracted tables and columns against our table and column index. Syntax errors can be fixed directly by passing the error back to a LLM, but hallucination often indicates that context lacked the required table or column to write the query. Researcher LLM Agent is equipped with tools to search for tables and retrieve their schemas and other metadata. For example, it can search for a table similar to the hallucinated one. It uses self-reflection to check its work and returns the updated context with a recommendation on what data to use \cite{NEURIPS2023_1b44b878}. To improve speed, we use a faster model (\texttt{gpt-4o-mini}) for the researcher LLM and a stronger model (\texttt{gpt-4o}) for self-reflection. The updated context and recommendation are sent to a query fixer LLM to fix the query. The architecture of the Researcher Agent is shown in Appendix~\ref{ref:researcher_agent_architecture}.

\subsection{User Interface and Multi-Agent Architecture}

In this section, we discuss how we deliver end-user utility through an interactive chat experience.

\subsubsection{Multi-Agent Architecture}

\begin{figure}[h]
    \vspace{-0.1in}
    \centering
    \includegraphics[width=0.9\linewidth]{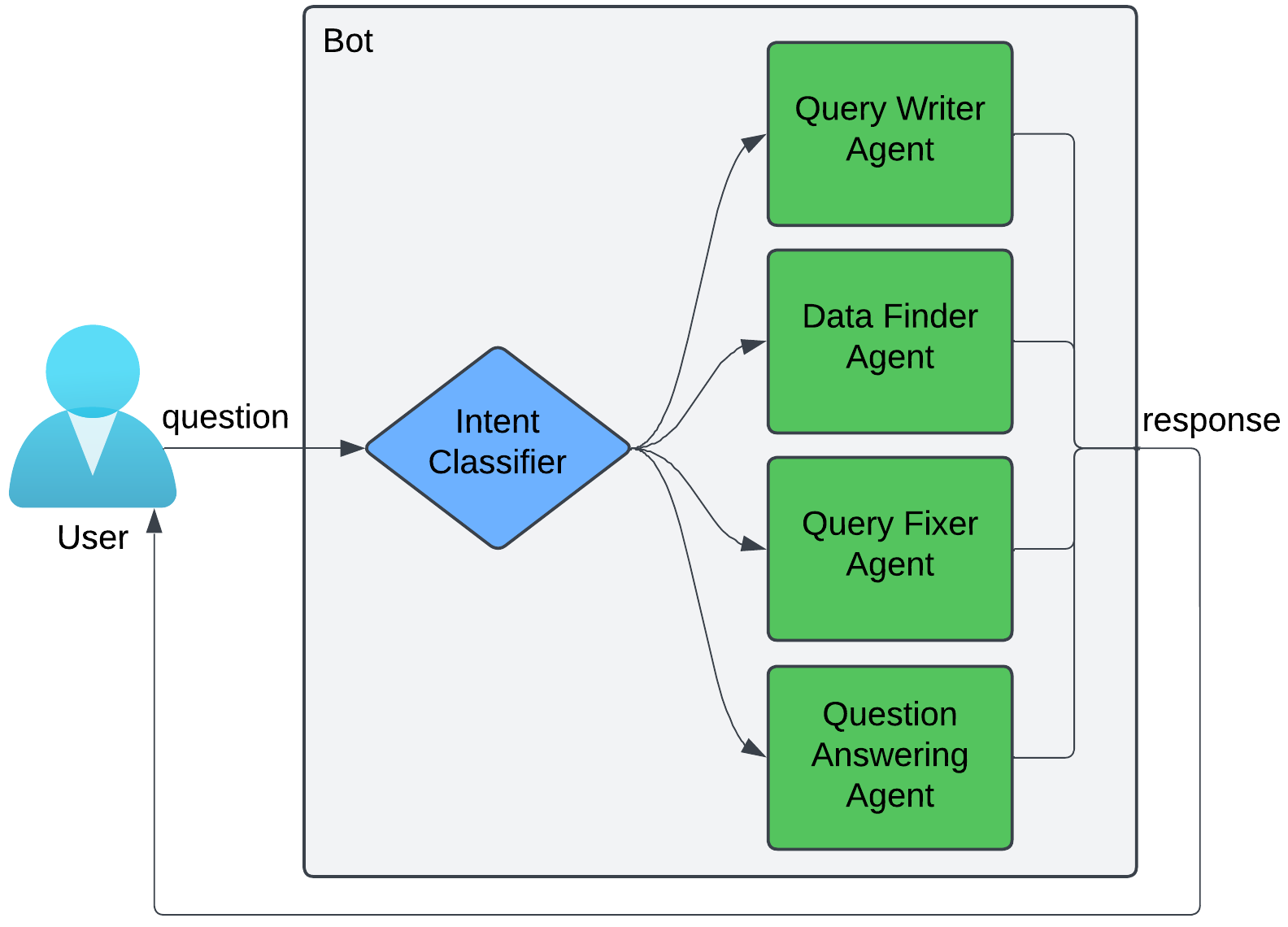}
    \caption{Multi-agent architecture supports various intents by routing questions to the appropriate agent.}
    \label{fig:model_architecture}
    \Description[User's question is routed to the appropriate agent]{The user's question is processed by an intent classifier and routed to one of four agents: Query Writer, Data Finder, Query Fixer, Question Answering Agent. The response is sent back to the user. This loop continues to support interactive chat.}
    \vspace{-0.1in}
\end{figure}

Post-launch, we learned that users want not only Text-to-SQL, but a general purpose data assistant that can answer questions about data, explain code, fix queries and handle follow-ups. Our multi-agent framework supports these questions with intent-specific agents to ensure high quality responses for the most common intents: query writing, data finding, and query fixing and another agent to answer the long-tail of other questions. We classify the user's intent and route the question to the appropriate agent, as shown in Figure~\ref{fig:model_architecture}. To answer follow-up questions quickly, agents maintain state, remembering context, progress, and conversation history. Agents can return different types of responses by calling only a subset of their tools (e.g. the Query Writer Agent can retrieve and rank context to return a list of suggested tables). The intent classifier classifies the follow-up intent to help the agent call the appropriate tools, routing unsupported intents to the Question-Answering Agent. The Question-Answering Agent has a similar architecture as the Researcher Agent. It is equipped with tools to get table schemas, search for tables, search wikis, and validate a query. To increase speed, it evaluates question difficulty, bypasses self-reflection for simple queries, and pre-fetches table metadata to minimize routine tool calls. It uses \texttt{gpt-4o-mini} for both LLMs.

\subsubsection{User Interface}
\label{sec:ui}

The chatbot's UI is designed to empower users by integrating it directly into the user's workflow. The chatbot can be accessed from the sidebar of our company's internal platform for SQL querying. In the chat area, users can ask questions and provide feedback (thumbs-up/down). After submitting a question, the user sees step-by-step progress updates from the bot. By default the product area is set based on the user's position in the company's organizational chart. Chatbot answers are presented to users using interactive display elements designed to educate and foster trust. \textbf{Query output} includes the generated query with inline comments, along with validation check results, query explanation, query tables, related reference queries, and assumptions to verify. \textbf{Table output} includes suggested tables along with their descriptions, popularity, commonly joined tables, and certification status. Checkboxes allow the user to select tables to use. Quick reply buttons above the chat box suggest follow-up options. Users can customize preferences in the UI by selecting product areas, adding domain knowledge, and certifying example queries. The user is shown a button if any query execution fails; this triggers the query fixer agent to debug errors related to data permissions, syntax, and schema correctness. An example of the chatbot's UI, table and query outputs, and one-click button to fix the query is shown in Appendix~\ref{ref:appendix_user_interface}.

\section{Evaluation}

\subsection{Benchmark Definition}
We defined an internal benchmark set to measure performance for our enterprise. We asked product area domain experts in our company to provide questions and ground truth queries. Since questions may have multiple valid solutions (e.g., using equivalent tables or columns), we accommodate multiple ground truth responses per question. In total, we collected 133 questions across 10 product areas and 167 ground truth tables. ~60\% of the questions have more than one valid response. These questions test different semantics within each product area so that we can evaluate our system's ability to understand the question, identify the right tables and columns, and generate semantically correct queries.

\subsection{Evaluation Metrics}
Execution accuracy (EX) is a popular evaluation metric used in public benchmarks such as Spider \cite{yu-etal-2018-spider}, BIRD \cite{li2023llmservedatabaseinterface}, and KaggleDBQA \cite{lee-etal-2021-kaggledbqa}. A SQL query is marked correct if its output matches the expected output. However, this does not account for other valid output formats or differentiate between queries that are almost correct (e.g. wrong date range) from those that are completely wrong.

Since our end user experience is an interactive chatbot, we focus on metrics that assess utility: did it find the right tables and columns, is the query correct or close to correct, and did it respond in a timely manner. We organize our evaluation metrics into these three categories. Recall metrics measure whether the bot found the correct tables and columns. Quality metrics measure overall accuracy (on a scale from 1-5), compilation success rate, and table / column hallucination rate. (Note that compilation success requires correct syntax, function usage, etc. in addition to valid tables and columns; we measure hallucination separately because it is a common issue with LLMs). Latency metrics include the average number of LLM calls, EBR queries, and database queries to generate the response. These are proxies for total runtime that give a sense of the complexity of the system. Speed is often overlooked in academic benchmarks, but is an important factor for practical deployment. In terms of actual runtime, our full model configuration runs in under 60 seconds per question on average.

We define table and column recall as the percentage of ground truth tables and columns present in the query, using the ground truth with the highest table overlap to the bot's response. For the overall score, we use both human evaluators and LLM-as-a-judge \cite{zheng2023judgingllmasajudgemtbenchchatbot} to rate queries. The inputs for evaluation are the user's question, bot response, ground truth queries, and rubric. A score of 5 is correct, 4 is close to correct, 3 identifies the right tables and most of the right columns; 2 has the wrong columns and 1 is completely wrong. Besides the overall score, we ask for correctness along various aspects (tables, columns, joins, filters, aggregations), query efficiency, and readability to identify aspects for improvement. The full scoring rubric is included in Appendix~\ref{ref:evaluation_rubric}.

\subsection{Results}
We ran ablation studies to measure the impact of each knowledge graph and modeling component on performance. For benchmarking, we used \texttt{E5-large-v2} \cite{wang2024textembeddingsweaklysupervisedcontrastive} for example embeddings and \texttt{text-embedding-ada-002} \cite{text_embedding_ada_002_open_ai} for table/column embeddings. We used \texttt{gpt-4o-mini-2024-07-18} \cite{open_ai_gpt_4o_mini} for the Researcher LLM and\\ 
\texttt{gpt-4o-2024-05-13} \cite{open_ai_gpt_4o} for the other LLMs, with temperature 0. We did a single run for each configuration. LLM-as-a-judge scoring is done with \texttt{gpt-4o-2024-05-13}. All models were accessed through the Azure OpenAI Service \cite{azure_open_ai_service}.
{
\setlength{\tabcolsep}{1.5pt}
\begin{table*}
\centering
\small
\begin{tabular}{
    p{0.05\linewidth}
    p{0.25\linewidth}
    >{\raggedleft\arraybackslash}p{0.08\linewidth}
    >{\raggedleft\arraybackslash}p{0.08\linewidth}
    >{\raggedleft\arraybackslash}p{0.08\linewidth}
    >{\raggedleft\arraybackslash}p{0.08\linewidth}
    >{\raggedleft\arraybackslash}p{0.08\linewidth}
    >{\raggedleft\arraybackslash}p{0.08\linewidth}
    >{\raggedleft\arraybackslash}p{0.08\linewidth}
    >{\raggedleft\arraybackslash}p{0.08\linewidth}
}
\toprule
\multicolumn{2}{c}{\textbf{Configuration}} & \multicolumn{2}{c}{\textbf{Recall Metrics}} & \multicolumn{3}{c}{\textbf{Quality Metrics}} & \multicolumn{3}{c}{\textbf{Latency Metrics}} \\
\cmidrule(lr){1-2} \cmidrule(lr){3-4} \cmidrule(lr){5-7} \cmidrule(lr){8-10}
\rule{0pt}{1.1em}Index & Description & Table Recall & Column Recall & Score (\% 4+) & \parbox[t]{\linewidth}{\raggedleft Successful compilation} & Valid tables \& columns & LLM calls & EBR queries & DB queries \\
\midrule
Full & All components & 78\% & 56\% & 48\% & 96\% & 99\% & 4.6 & 3.0 & 9.4 \\
\hline
A.5 & Full w/o popularity or joins & 77\% & 53\% & 42\% & 95\% & 98\% & 4.8 & 3.0 & 8.4 \\
A.4 & A.5 w/o domain knowledge or jargon & 76\% & 52\% & 49\% & 96\% & 99\% & 4.7 & 3.0 & 8.5 \\
A.3 & A.4 w/o example queries & 60\% & 38\% & 24\% & 98\% & 100\% & 4.6 & 1.0 & 7.0 \\
A.2 & A.3 w/o table or column attributes & 56\% & 30\% & 11\% & 93\% & 99\% & 5.0 & 1.0 & 7.5 \\
A.1 & A.2 w/o table clusters (schemas only) & 45\% & 24\% & 9\% & 88\% & 99\% & 5.1 & 1.0 & 7.1 \\
\hline
B.3 & Full w/o researcher agent & 75\% & 53\% & 47\% & 95\% & 98\% & 4.3 & 3.0 & 9.5 \\
B.2 & B.3 w/o query fixer & 76\% & 55\% & 47\% & 76\% & 85\% & 4.0 & 3.0 & 8.4 \\
B.1 & B.2 w/o rankers (EBR, writer only) & 67\% & 50\% & 46\% & 66\% & 77\% & 2.0 & 3.0 & 7.1 \\
\hline
C.4 & (A.4, B.3) combination & 76\% & 52\% & 46\% & 97\% & 98\% & 4.3 & 3.0 & 8.6 \\
C.3 & (A.3, B.2) combination & 60\% & 37\% & 20\% & 77\% & 87\% & 4.0 & 1.0 & 6.1 \\
C.2 & (A.2, B.1) combination  & 49\% & 27\% & 17\% & 68\% & 83\% & 2.0 & 1.0 & 5.0 \\
C.1 & (A.1, B.1) combination & 37\% & 23\% & 16\% & 67\% & 81\% & 1.9 & 1.0 & 3.9 \\
\bottomrule

\end{tabular}
\caption{Ablation study. A.1-A.5 show the effect of ablating knowledge graph components. B.1-B.3 show the effect of ablating modeling components. C.1-C.4 show the effect of ablating both.}
\label{tab:ablation}
\end{table*}
}

\subsubsection{Ablation of Knowledge Graph Components}
We keep all the modeling components and remove knowledge graph components in sequence, from hardest to easiest to collect. The simplest configuration uses only table schemas. When table descriptions are removed, we embed the table schema. Results are shown in Table~\ref{tab:ablation} (A.1-A.5). Knowledge graph components improve the \% of answers scoring 4+ (correct or close to correct) from 9\% (A.1, schemas only) to 48\% (`Full'). Most of the improvement is from example queries, table clustering, and table/column attributes (i.e. description and other attributes from Table~\ref{tab:knowledge_graph}, besides schema and popularity). Surprisingly, adding domain knowledge decreases the percentage of answers scoring 4+; perhaps irrelevant domain knowledge records lower performance.

\subsubsection{Ablation of Modeling Components}
We keep all the knowledge graph components and remove modeling components in order from most to least complex. The simplest configuration uses only EBR and Query Writer. When the researcher agent is disabled, the query fixer is given schemas for the query tables along with tables after ranking that contain hallucinated columns. When the context rankers are disabled, we pass the first $K_{\text{table}}=7$ tables from EBR and all their columns directly to the query writer. Results are shown in Table~\ref{tab:ablation} (B.1-B.3). Modeling components improve compilation success rate from 66\% (B.1) to 96\% (`Full') and valid tables \& columns (no schema hallucination) from 77\% to 99\%, largely from the query fixer and context rankers. The context rankers also improve recall. Interestingly, there are minimal improvements in score. This suggests that generating a correct or close to correct query requires understanding data semantics. The knowledge graph is a better lever to improve semantic understanding, although it does require adequate utilization by modeling components.

\subsubsection{Ablation of Both Components}
\label{subsubsec:ablation_both}
We are also interested in configurations that use a subset of knowledge graph and modeling components in the order a developer might build a Text-to-SQL solution. We benchmark four configurations from small to large. Results are shown in Table~\ref{tab:ablation} (C.1-C.4). Compared to C.1, configurations C.2 and C.3 improve recall and error rates, but quality does not improve much until C.4. Therefore, many of the components described in this paper were necessary to provide a good experience for our end users. C.4 has similar performance to the full model while requiring fewer LLM calls and database queries.

\subsubsection{Expert Review}
We asked human experts from each product area to evaluate the chatbot's answers on the production version of our model (slightly modified from the `Full' configuration). We received ratings for 124 of 133 questions. 53\% were rated 4 or 5 (minor issues or correct). 77\% were rated 3+ (helpful for identifying tables and columns). The most common issue was `Filter is incorrect' (24\%). On the other hand, only 4\% had an incorrect join.

\section{Deployment}
Our chatbot has been available in the querying platform since July 2024 with steady usage of over 300 weekly active users. Our most popular intents are query fixing, query writing, and data discovery. Power users have over 100 chat sessions / month. About 20\% of weekly active users return the following week. 33\% of chat sessions lead to code pasted from the chatbot into the SQL editor. We conducted a survey to understand users' satisfaction with the chatbot. 39\% of users rated its queries as ``Very good (require minor modifications)'' or ``Excellent (queries are correct).'' 95\% of users rated them at least ``Passes (queries require some modifications).'' Our knowledge graph allows users to improve their experience by defining product areas, adding example queries, table / column documentation, and domain knowledge.

\section{Conclusion}
We have demonstrated a successful approach to building a Text-to-SQL chatbot for enterprise data analytics. To incorporate semantic understanding, we constructed a knowledge graph that includes information derived from the database and from enterprise users. Incorporating example queries, table clusters, and table / column attributes improved the percentage of correct or almost correct queries from 9\% to 49\% compared to using table schemas alone. Through modeling components such as retrieval, ranking, and query-fixing with researcher agent, we achieved 78\% table recall and reduced schema hallucination from 23\% to 1\%, compilation errors from 34\% to 4\%. Finally, we showed how we built an application that helps non-experts write SQL queries through integration into the SQL editor, rich display elements, and ability to support different intents and follow-up chats through a multi-agent framework. We hope these insights will accelerate the development of industry solutions in this space.

\begin{acks}
    We would like to thank the Data Science team for curating the benchmark dataset and evaluating query accuracy,
    and Erik Krogen and Slim Bouguerra for assistance with the Trino query plan parser.
\end{acks}
\bibliographystyle{ACM-Reference-Format}
\bibliography{custom}


\begin{thebibliography}{35}


\ifx \showCODEN    \undefined \def \showCODEN     #1{\unskip}     \fi
\ifx \showISBNx    \undefined \def \showISBNx     #1{\unskip}     \fi
\ifx \showISBNxiii \undefined \def \showISBNxiii  #1{\unskip}     \fi
\ifx \showISSN     \undefined \def \showISSN      #1{\unskip}     \fi
\ifx \showLCCN     \undefined \def \showLCCN      #1{\unskip}     \fi
\ifx \shownote     \undefined \def \shownote      #1{#1}          \fi
\ifx \showarticletitle \undefined \def \showarticletitle #1{#1}   \fi
\ifx \showURL      \undefined \def \showURL       {\relax}        \fi
\providecommand\bibfield[2]{#2}
\providecommand\bibinfo[2]{#2}
\providecommand\natexlab[1]{#1}
\providecommand\showeprint[2][]{arXiv:#2}

\bibitem[Claveau(2021)]%
        {claveau2021neural}
\bibfield{author}{\bibinfo{person}{Vincent Claveau}.} \bibinfo{year}{2021}\natexlab{}.
\newblock \showarticletitle{Neural text generation for query expansion in information retrieval}. In \bibinfo{booktitle}{\emph{IEEE/WIC/ACM International Conference on Web Intelligence and Intelligent Agent Technology}}. \bibinfo{pages}{202--209}.
\newblock


\bibitem[{datahub}-{project}(2024)]%
        {datahub}
\bibfield{author}{\bibinfo{person}{{datahub}-{project}}.} \bibinfo{year}{2024}\natexlab{}.
\newblock \bibinfo{title}{DataHub}.
\newblock \bibinfo{howpublished}{\url{https://github.com/datahub-project/datahub}}.
\newblock


\bibitem[Floratou et~al\mbox{.}(2024)]%
        {floratou_nl2sql_2024}
\bibfield{author}{\bibinfo{person}{Avrilia Floratou}, \bibinfo{person}{Fotis Psallidas}, \bibinfo{person}{Fuheng Zhao}, \bibinfo{person}{Shaleen Deep}, \bibinfo{person}{Gunther Hagleither}, \bibinfo{person}{Wangda Tan}, \bibinfo{person}{Joyce Cahoon}, \bibinfo{person}{Rana Alotaibi}, \bibinfo{person}{Jordan Henkel}, \bibinfo{person}{Abhik Singla}, \bibinfo{person}{Alex~Van Grootel}, \bibinfo{person}{Brandon Chow}, \bibinfo{person}{Kai Deng}, \bibinfo{person}{Katherine Lin}, \bibinfo{person}{Marcos Campos}, \bibinfo{person}{K.~Venkatesh Emani}, \bibinfo{person}{Vivek Pandit}, \bibinfo{person}{Victor Shnayder}, \bibinfo{person}{Wenjing Wang}, {and} \bibinfo{person}{Carlo Curino}.} \bibinfo{year}{2024}\natexlab{}.
\newblock \showarticletitle{NL2SQL is a solved problem... Not!}. In \bibinfo{booktitle}{\emph{Conference on Innovative Data Systems Research}}.
\newblock
\urldef\tempurl%
\url{https://api.semanticscholar.org/CorpusID:266729311}
\showURL{%
\tempurl}


\bibitem[Foundation(2025a)]%
        {trino_explain}
\bibfield{author}{\bibinfo{person}{Trino~Software Foundation}.} \bibinfo{year}{2025}\natexlab{a}.
\newblock \bibinfo{title}{EXPLAIN}.
\newblock \bibinfo{howpublished}{\url{https://trino.io/docs/current/sql/explain.html}}.
\newblock
\newblock
\shownote{Accessed: 2025-04-25}.


\bibitem[Foundation(2025b)]%
        {trino}
\bibfield{author}{\bibinfo{person}{Trino~Software Foundation}.} \bibinfo{year}{2025}\natexlab{b}.
\newblock \bibinfo{title}{Trino: The Distributed SQL Query Engine}.
\newblock \bibinfo{howpublished}{\url{https://trino.io}}.
\newblock
\newblock
\shownote{Version 475}.


\bibitem[Gao et~al\mbox{.}(2023)]%
        {gao_text--sql_2023}
\bibfield{author}{\bibinfo{person}{Dawei Gao}, \bibinfo{person}{Haibin Wang}, \bibinfo{person}{Yaliang Li}, \bibinfo{person}{Xiuyu Sun}, \bibinfo{person}{Yichen Qian}, \bibinfo{person}{Bolin Ding}, {and} \bibinfo{person}{Jingren Zhou}.} \bibinfo{year}{2023}\natexlab{}.
\newblock \bibinfo{title}{Text-to-{SQL} {Empowered} by {Large} {Language} {Models}: {A} {Benchmark} {Evaluation}}.
\newblock
\urldef\tempurl%
\url{http://arxiv.org/abs/2308.15363}
\showURL{%
\tempurl}
\newblock
\shownote{arXiv:2308.15363 [cs]}.


\bibitem[Glean(2024)]%
        {glean_chat_api}
\bibfield{author}{\bibinfo{person}{Glean}.} \bibinfo{year}{2024}\natexlab{}.
\newblock \bibinfo{title}{Glean Chat API}.
\newblock
\urldef\tempurl%
\url{https://developers.glean.com/docs/client_api/chat_api/}
\showURL{%
\tempurl}


\bibitem[Hyvärinen and Oja(2000)]%
        {HYVARINEN2000411}
\bibfield{author}{\bibinfo{person}{Aapo Hyvärinen} {and} \bibinfo{person}{Erkki Oja}.} \bibinfo{year}{2000}\natexlab{}.
\newblock \showarticletitle{Independent component analysis: algorithms and applications}.
\newblock \bibinfo{journal}{\emph{Neural Networks}} \bibinfo{volume}{13}, \bibinfo{number}{4} (\bibinfo{year}{2000}), \bibinfo{pages}{411--430}.
\newblock
\showISSN{0893-6080}
\href{https://doi.org/10.1016/S0893-6080(00)00026-5}{doi:\nolinkurl{10.1016/S0893-6080(00)00026-5}}


\bibitem[Jagerman et~al\mbox{.}(2023)]%
        {jagerman2023queryexpansionpromptinglarge}
\bibfield{author}{\bibinfo{person}{Rolf Jagerman}, \bibinfo{person}{Honglei Zhuang}, \bibinfo{person}{Zhen Qin}, \bibinfo{person}{Xuanhui Wang}, {and} \bibinfo{person}{Michael Bendersky}.} \bibinfo{year}{2023}\natexlab{}.
\newblock \bibinfo{title}{Query Expansion by Prompting Large Language Models}.
\newblock
\showeprint[arxiv]{2305.03653}~[cs.IR]
\urldef\tempurl%
\url{https://arxiv.org/abs/2305.03653}
\showURL{%
\tempurl}


\bibitem[Khune et~al\mbox{.}(2024)]%
        {query_gpt_uber}
\bibfield{author}{\bibinfo{person}{Abhi Khune}, \bibinfo{person}{Callie Busch}, \bibinfo{person}{Jeffrey Johnson}, \bibinfo{person}{Pradeep Chakka}, \bibinfo{person}{Saketh Chintapalli}, \bibinfo{person}{Adarsh Nagesh}, \bibinfo{person}{Gaurav Paul}, {and} \bibinfo{person}{Ben Carroll}.} \bibinfo{year}{2024}\natexlab{}.
\newblock \bibinfo{title}{QueryGPT - Natural Language to SQL Using Generative AI}.
\newblock
\urldef\tempurl%
\url{https://www.uber.com/blog/query-gpt/}
\showURL{%
\tempurl}


\bibitem[LangChain(2024)]%
        {langchain}
\bibfield{author}{\bibinfo{person}{LangChain}.} \bibinfo{year}{2024}\natexlab{}.
\newblock \bibinfo{booktitle}{\emph{LangChain}}.
\newblock
\urldef\tempurl%
\url{https://www.langchain.com/}
\showURL{%
\tempurl}


\bibitem[Lee et~al\mbox{.}(2021)]%
        {lee-etal-2021-kaggledbqa}
\bibfield{author}{\bibinfo{person}{Chia-Hsuan Lee}, \bibinfo{person}{Oleksandr Polozov}, {and} \bibinfo{person}{Matthew Richardson}.} \bibinfo{year}{2021}\natexlab{}.
\newblock \showarticletitle{{K}aggle{DBQA}: Realistic Evaluation of Text-to-{SQL} Parsers}. In \bibinfo{booktitle}{\emph{ACL:2021:long}}, \bibfield{editor}{\bibinfo{person}{Chengqing Zong}, \bibinfo{person}{Fei Xia}, \bibinfo{person}{Wenjie Li}, {and} \bibinfo{person}{Roberto Navigli}} (Eds.). \bibinfo{address}{Online}, \bibinfo{pages}{2261--2273}.
\newblock
\href{https://doi.org/10.18653/v1/2021.acl-long.176}{doi:\nolinkurl{10.18653/v1/2021.acl-long.176}}


\bibitem[Lee et~al\mbox{.}(2024)]%
        {lee2024mcs}
\bibfield{author}{\bibinfo{person}{Dongjun Lee}, \bibinfo{person}{Choongwon Park}, \bibinfo{person}{Jaehyuk Kim}, {and} \bibinfo{person}{Heesoo Park}.} \bibinfo{year}{2024}\natexlab{}.
\newblock \showarticletitle{Mcs-sql: Leveraging multiple prompts and multiple-choice selection for text-to-sql generation}.
\newblock \bibinfo{journal}{\emph{arXiv preprint arXiv:2405.07467}} (\bibinfo{year}{2024}).
\newblock


\bibitem[Lei et~al\mbox{.}(2024)]%
        {lei2024spider20evaluatinglanguage}
\bibfield{author}{\bibinfo{person}{Fangyu Lei}, \bibinfo{person}{Jixuan Chen}, \bibinfo{person}{Yuxiao Ye}, \bibinfo{person}{Ruisheng Cao}, \bibinfo{person}{Dongchan Shin}, \bibinfo{person}{Hongjin Su}, \bibinfo{person}{Zhaoqing Suo}, \bibinfo{person}{Hongcheng Gao}, \bibinfo{person}{Wenjing Hu}, \bibinfo{person}{Pengcheng Yin}, \bibinfo{person}{Victor Zhong}, \bibinfo{person}{Caiming Xiong}, \bibinfo{person}{Ruoxi Sun}, \bibinfo{person}{Qian Liu}, \bibinfo{person}{Sida Wang}, {and} \bibinfo{person}{Tao Yu}.} \bibinfo{year}{2024}\natexlab{}.
\newblock \bibinfo{title}{Spider 2.0: Evaluating Language Models on Real-World Enterprise Text-to-SQL Workflows}.
\newblock
\showeprint[arxiv]{2411.07763}~[cs.CL]
\urldef\tempurl%
\url{https://arxiv.org/abs/2411.07763}
\showURL{%
\tempurl}


\bibitem[Li et~al\mbox{.}(2023)]%
        {li2023llmservedatabaseinterface}
\bibfield{author}{\bibinfo{person}{Jinyang Li}, \bibinfo{person}{Binyuan Hui}, \bibinfo{person}{Ge Qu}, \bibinfo{person}{Jiaxi Yang}, \bibinfo{person}{Binhua Li}, \bibinfo{person}{Bowen Li}, \bibinfo{person}{Bailin Wang}, \bibinfo{person}{Bowen Qin}, \bibinfo{person}{Rongyu Cao}, \bibinfo{person}{Ruiying Geng}, \bibinfo{person}{Nan Huo}, \bibinfo{person}{Xuanhe Zhou}, \bibinfo{person}{Chenhao Ma}, \bibinfo{person}{Guoliang Li}, \bibinfo{person}{Kevin C.~C. Chang}, \bibinfo{person}{Fei Huang}, \bibinfo{person}{Reynold Cheng}, {and} \bibinfo{person}{Yongbin Li}.} \bibinfo{year}{2023}\natexlab{}.
\newblock \bibinfo{title}{Can LLM Already Serve as A Database Interface? A BIg Bench for Large-Scale Database Grounded Text-to-SQLs}.
\newblock
\showeprint[arxiv]{2305.03111}~[cs.CL]
\urldef\tempurl%
\url{https://arxiv.org/abs/2305.03111}
\showURL{%
\tempurl}


\bibitem[Maamari et~al\mbox{.}(2024)]%
        {maamari_death_2024}
\bibfield{author}{\bibinfo{person}{Karime Maamari}, \bibinfo{person}{Fadhil Abubaker}, \bibinfo{person}{Daniel Jaroslawicz}, {and} \bibinfo{person}{Amine Mhedhbi}.} \bibinfo{year}{2024}\natexlab{}.
\newblock \bibinfo{title}{The {Death} of {Schema} {Linking}? {Text}-to-{SQL} in the {Age} of {Well}-{Reasoned} {Language} {Models}}.
\newblock
\urldef\tempurl%
\url{http://arxiv.org/abs/2408.07702}
\showURL{%
\tempurl}
\newblock
\shownote{arXiv:2408.07702}.


\bibitem[Microsoft(2025)]%
        {azure_open_ai_service}
\bibfield{author}{\bibinfo{person}{Microsoft}.} \bibinfo{year}{2025}\natexlab{}.
\newblock \bibinfo{title}{Azure OpenAI in Azure AI Foundry Models}.
\newblock
\urldef\tempurl%
\url{https://learn.microsoft.com/en-us/azure/ai-foundry/openai/concepts/models}
\showURL{%
\tempurl}


\bibitem[OpenAI(2022)]%
        {text_embedding_ada_002_open_ai}
\bibfield{author}{\bibinfo{person}{OpenAI}.} \bibinfo{year}{2022}\natexlab{}.
\newblock \bibinfo{title}{text-embedding-ada-002}.
\newblock
\urldef\tempurl%
\url{https://platform.openai.com/docs/models/text-embedding-ada-002}
\showURL{%
\tempurl}


\bibitem[OpenAI(2023)]%
        {open_ai_gpt_35_turbo}
\bibfield{author}{\bibinfo{person}{OpenAI}.} \bibinfo{year}{2023}\natexlab{}.
\newblock \bibinfo{title}{GPT-3.5 Turbo}.
\newblock
\urldef\tempurl%
\url{https://platform.openai.com/docs/models/gpt-3.5-turbo}
\showURL{%
\tempurl}


\bibitem[OpenAI(2024a)]%
        {open_ai_gpt_4o}
\bibfield{author}{\bibinfo{person}{OpenAI}.} \bibinfo{year}{2024}\natexlab{a}.
\newblock \bibinfo{title}{GPT-4o}.
\newblock
\urldef\tempurl%
\url{https://platform.openai.com/docs/models/gpt-4o}
\showURL{%
\tempurl}


\bibitem[OpenAI(2024b)]%
        {open_ai_gpt_4o_mini}
\bibfield{author}{\bibinfo{person}{OpenAI}.} \bibinfo{year}{2024}\natexlab{b}.
\newblock \bibinfo{title}{GPT-4o mini}.
\newblock
\urldef\tempurl%
\url{https://platform.openai.com/docs/models/gpt-4o-mini}
\showURL{%
\tempurl}


\bibitem[Pedregosa et~al\mbox{.}(2011)]%
        {pedregosa2011scikit}
\bibfield{author}{\bibinfo{person}{Fabian Pedregosa}, \bibinfo{person}{Ga{\"e}l Varoquaux}, \bibinfo{person}{Alexandre Gramfort}, \bibinfo{person}{Vincent Michel}, \bibinfo{person}{Bertrand Thirion}, \bibinfo{person}{Olivier Grisel}, \bibinfo{person}{Mathieu Blondel}, \bibinfo{person}{Peter Prettenhofer}, \bibinfo{person}{Ron Weiss}, \bibinfo{person}{Vincent Dubourg}, {et~al\mbox{.}}} \bibinfo{year}{2011}\natexlab{}.
\newblock \showarticletitle{Scikit-learn: Machine learning in Python}.
\newblock \bibinfo{journal}{\emph{Journal of machine learning research}} \bibinfo{volume}{12}, \bibinfo{number}{Oct} (\bibinfo{year}{2011}), \bibinfo{pages}{2825--2830}.
\newblock


\bibitem[Pourreza et~al\mbox{.}(2024)]%
        {pourreza_chase-sql_2024}
\bibfield{author}{\bibinfo{person}{Mohammadreza Pourreza}, \bibinfo{person}{Hailong Li}, \bibinfo{person}{Ruoxi Sun}, \bibinfo{person}{Yeounoh Chung}, \bibinfo{person}{Shayan Talaei}, \bibinfo{person}{Gaurav~Tarlok Kakkar}, \bibinfo{person}{Yu Gan}, \bibinfo{person}{Amin Saberi}, \bibinfo{person}{Fatma Ozcan}, {and} \bibinfo{person}{Sercan~O. Arik}.} \bibinfo{year}{2024}\natexlab{}.
\newblock \bibinfo{title}{{CHASE}-{SQL}: {Multi}-{Path} {Reasoning} and {Preference} {Optimized} {Candidate} {Selection} in {Text}-to-{SQL}}.
\newblock
\urldef\tempurl%
\url{http://arxiv.org/abs/2410.01943}
\showURL{%
\tempurl}
\newblock
\shownote{arXiv:2410.01943}.


\bibitem[Pourreza and Rafiei(2023)]%
        {pourreza_din-sql_2023}
\bibfield{author}{\bibinfo{person}{Mohammadreza Pourreza} {and} \bibinfo{person}{Davood Rafiei}.} \bibinfo{year}{2023}\natexlab{}.
\newblock \bibinfo{title}{{DIN}-{SQL}: {Decomposed} {In}-{Context} {Learning} of {Text}-to-{SQL} with {Self}-{Correction}}.
\newblock
\href{https://doi.org/10.48550/arXiv.2304.11015}{doi:\nolinkurl{10.48550/arXiv.2304.11015}}
\newblock
\shownote{arXiv:2304.11015 [cs]}.


\bibitem[Shinn et~al\mbox{.}(2023)]%
        {NEURIPS2023_1b44b878}
\bibfield{author}{\bibinfo{person}{Noah Shinn}, \bibinfo{person}{Federico Cassano}, \bibinfo{person}{Ashwin Gopinath}, \bibinfo{person}{Karthik Narasimhan}, {and} \bibinfo{person}{Shunyu Yao}.} \bibinfo{year}{2023}\natexlab{}.
\newblock \showarticletitle{Reflexion: language agents with verbal reinforcement learning}. In \bibinfo{booktitle}{\emph{Advances in Neural Information Processing Systems}}, \bibfield{editor}{\bibinfo{person}{A.~Oh}, \bibinfo{person}{T.~Naumann}, \bibinfo{person}{A.~Globerson}, \bibinfo{person}{K.~Saenko}, \bibinfo{person}{M.~Hardt}, {and} \bibinfo{person}{S.~Levine}} (Eds.), Vol.~\bibinfo{volume}{36}. \bibinfo{publisher}{Curran Associates, Inc.}, \bibinfo{pages}{8634--8652}.
\newblock
\urldef\tempurl%
\url{https://proceedings.neurips.cc/paper_files/paper/2023/file/1b44b878bb782e6954cd888628510e90-Paper-Conference.pdf}
\showURL{%
\tempurl}


\bibitem[Talaei et~al\mbox{.}(2024)]%
        {talaei_chess_2024}
\bibfield{author}{\bibinfo{person}{Shayan Talaei}, \bibinfo{person}{Mohammadreza Pourreza}, \bibinfo{person}{Yu-Chen Chang}, \bibinfo{person}{Azalia Mirhoseini}, {and} \bibinfo{person}{Amin Saberi}.} \bibinfo{year}{2024}\natexlab{}.
\newblock \bibinfo{title}{{CHESS}: {Contextual} {Harnessing} for {Efficient} {SQL} {Synthesis}}.
\newblock
\urldef\tempurl%
\url{http://arxiv.org/abs/2405.16755}
\showURL{%
\tempurl}
\newblock
\shownote{arXiv:2405.16755 [cs]}.


\bibitem[Team(2025a)]%
        {bird_leaderboard}
\bibfield{author}{\bibinfo{person}{BIRD Team}.} \bibinfo{year}{2025}\natexlab{a}.
\newblock \bibinfo{title}{BIRD-SQL: A Big Bench for Large-Scale Database Grounded Text-to-SQLs}.
\newblock \bibinfo{howpublished}{\url{https://bird-benchmark.github.io/}}.
\newblock
\newblock
\shownote{Accessed: 2025-04-25}.


\bibitem[Team(2025b)]%
        {spider_leaderboard}
\bibfield{author}{\bibinfo{person}{Spider Team}.} \bibinfo{year}{2025}\natexlab{b}.
\newblock \bibinfo{title}{Spider 1.0: Yale Semantic Parsing and Text-to-SQL Challenge}.
\newblock \bibinfo{howpublished}{\url{https://yale-lily.github.io/spider}}.
\newblock
\newblock
\shownote{Accessed: 2025-04-25}.


\bibitem[Team(2025c)]%
        {spider2_leaderboard}
\bibfield{author}{\bibinfo{person}{Spider Team}.} \bibinfo{year}{2025}\natexlab{c}.
\newblock \bibinfo{title}{Spider 2.0: Evaluating Language Models on Real-World Enterprise Text-to-SQL Workflows}.
\newblock \bibinfo{howpublished}{\url{https://spider2-sql.github.io/}}.
\newblock
\newblock
\shownote{Accessed: 2025-04-25}.


\bibitem[Wang et~al\mbox{.}(2024)]%
        {wang2024textembeddingsweaklysupervisedcontrastive}
\bibfield{author}{\bibinfo{person}{Liang Wang}, \bibinfo{person}{Nan Yang}, \bibinfo{person}{Xiaolong Huang}, \bibinfo{person}{Binxing Jiao}, \bibinfo{person}{Linjun Yang}, \bibinfo{person}{Daxin Jiang}, \bibinfo{person}{Rangan Majumder}, {and} \bibinfo{person}{Furu Wei}.} \bibinfo{year}{2024}\natexlab{}.
\newblock \bibinfo{title}{Text Embeddings by Weakly-Supervised Contrastive Pre-training}.
\newblock
\showeprint[arxiv]{2212.03533}~[cs.CL]
\urldef\tempurl%
\url{https://arxiv.org/abs/2212.03533}
\showURL{%
\tempurl}


\bibitem[Wang et~al\mbox{.}(2023)]%
        {wang-etal-2023-query2doc}
\bibfield{author}{\bibinfo{person}{Liang Wang}, \bibinfo{person}{Nan Yang}, {and} \bibinfo{person}{Furu Wei}.} \bibinfo{year}{2023}\natexlab{}.
\newblock \showarticletitle{Query2doc: Query Expansion with Large Language Models}. In \bibinfo{booktitle}{\emph{EMNLP:2023:main}}, \bibfield{editor}{\bibinfo{person}{Houda Bouamor}, \bibinfo{person}{Juan Pino}, {and} \bibinfo{person}{Kalika Bali}} (Eds.). \bibinfo{address}{Singapore}, \bibinfo{pages}{9414--9423}.
\newblock
\href{https://doi.org/10.18653/v1/2023.emnlp-main.585}{doi:\nolinkurl{10.18653/v1/2023.emnlp-main.585}}


\bibitem[Wei et~al\mbox{.}(2022)]%
        {wei2022chain}
\bibfield{author}{\bibinfo{person}{Jason Wei}, \bibinfo{person}{Xuezhi Wang}, \bibinfo{person}{Dale Schuurmans}, \bibinfo{person}{Maarten Bosma}, \bibinfo{person}{Fei Xia}, \bibinfo{person}{Ed Chi}, \bibinfo{person}{Quoc~V Le}, \bibinfo{person}{Denny Zhou}, {et~al\mbox{.}}} \bibinfo{year}{2022}\natexlab{}.
\newblock \showarticletitle{Chain-of-thought prompting elicits reasoning in large language models}.
\newblock \bibinfo{journal}{\emph{Advances in neural information processing systems}}  \bibinfo{volume}{35} (\bibinfo{year}{2022}), \bibinfo{pages}{24824--24837}.
\newblock


\bibitem[Yu et~al\mbox{.}(2018)]%
        {yu-etal-2018-spider}
\bibfield{author}{\bibinfo{person}{Tao Yu}, \bibinfo{person}{Rui Zhang}, \bibinfo{person}{Kai Yang}, \bibinfo{person}{Michihiro Yasunaga}, \bibinfo{person}{Dongxu Wang}, \bibinfo{person}{Zifan Li}, \bibinfo{person}{James Ma}, \bibinfo{person}{Irene Li}, \bibinfo{person}{Qingning Yao}, \bibinfo{person}{Shanelle Roman}, \bibinfo{person}{Zilin Zhang}, {and} \bibinfo{person}{Dragomir Radev}.} \bibinfo{year}{2018}\natexlab{}.
\newblock \showarticletitle{{S}pider: A Large-Scale Human-Labeled Dataset for Complex and Cross-Domain Semantic Parsing and Text-to-{SQL} Task}. In \bibinfo{booktitle}{\emph{EMNLP:2018:1}}, \bibfield{editor}{\bibinfo{person}{Ellen Riloff}, \bibinfo{person}{David Chiang}, \bibinfo{person}{Julia Hockenmaier}, {and} \bibinfo{person}{Jun{'}ichi Tsujii}} (Eds.). \bibinfo{address}{Brussels, Belgium}, \bibinfo{pages}{3911--3921}.
\newblock
\href{https://doi.org/10.18653/v1/D18-1425}{doi:\nolinkurl{10.18653/v1/D18-1425}}


\bibitem[Zheng et~al\mbox{.}(2023)]%
        {zheng2023judgingllmasajudgemtbenchchatbot}
\bibfield{author}{\bibinfo{person}{Lianmin Zheng}, \bibinfo{person}{Wei-Lin Chiang}, \bibinfo{person}{Ying Sheng}, \bibinfo{person}{Siyuan Zhuang}, \bibinfo{person}{Zhanghao Wu}, \bibinfo{person}{Yonghao Zhuang}, \bibinfo{person}{Zi Lin}, \bibinfo{person}{Zhuohan Li}, \bibinfo{person}{Dacheng Li}, \bibinfo{person}{Eric~P. Xing}, \bibinfo{person}{Hao Zhang}, \bibinfo{person}{Joseph~E. Gonzalez}, {and} \bibinfo{person}{Ion Stoica}.} \bibinfo{year}{2023}\natexlab{}.
\newblock \bibinfo{title}{Judging LLM-as-a-Judge with MT-Bench and Chatbot Arena}.
\newblock
\showeprint[arxiv]{2306.05685}~[cs.CL]
\urldef\tempurl%
\url{https://arxiv.org/abs/2306.05685}
\showURL{%
\tempurl}


\bibitem[Zhong et~al\mbox{.}(2020)]%
        {zhong-etal-2020-grounded}
\bibfield{author}{\bibinfo{person}{Victor Zhong}, \bibinfo{person}{Mike Lewis}, \bibinfo{person}{Sida~I. Wang}, {and} \bibinfo{person}{Luke Zettlemoyer}.} \bibinfo{year}{2020}\natexlab{}.
\newblock \showarticletitle{Grounded Adaptation for Zero-shot Executable Semantic Parsing}. In \bibinfo{booktitle}{\emph{EMNLP:2020:main}}, \bibfield{editor}{\bibinfo{person}{Bonnie Webber}, \bibinfo{person}{Trevor Cohn}, \bibinfo{person}{Yulan He}, {and} \bibinfo{person}{Yang Liu}} (Eds.). \bibinfo{address}{Online}, \bibinfo{pages}{6869--6882}.
\newblock
\href{https://doi.org/10.18653/v1/2020.emnlp-main.558}{doi:\nolinkurl{10.18653/v1/2020.emnlp-main.558}}


\end{thebibliography}
\newpage
\appendix
\section{Appendix}

\subsection{Evaluation Rubric}
\label{ref:evaluation_rubric}
Fig~\ref{fig:rubric} shows the overall scoring rubric for human evaluation and LLM-as-a-judge. In addition to this overall score, we ask reviewers if any table, column, filter, aggregation, join, etc. is incorrect.

\begin{figure}[H]
\centering
\lstset{
    frame=single,
    numbers=none,
    basicstyle=\ttfamily\small,
    columns=flexible,
    breaklines=true,
    breakatwhitespace=true,
}
\begin{lstlisting}
Overall rating of the query between 1-5 where higher score indicates higher quality. 

1 - The query is completely wrong and does not answer user's question at all.
2 - The query found the right tables but 90% of the columns are wrong and do not answer the user's question at all.
3 - The query has the right tables and majority of the right columns but has gaps that require substantial effort or domain knowledge to detect/fix and does not answer the user's question.
4 - The query has the right tables and almost all right columns but may have minor issues with the logic that are easy for a non-expert to fix. The query answers the user's question but may be missing some trivial details, e.g. filtering by date is incorrect.
5 - The query answers the user's question perfectly and answers the user's question completely and correctly.
\end{lstlisting}
\caption{Scoring rubric for evaluation.}
\label{fig:rubric}
\end{figure}

\subsection{User-Dataset Clustering}
\label{ref:clustering_pseudocode}
We provide the clustering algorithm to identify relevant datasets for each group of users. Algorithm~\ref{alg:cluster_datasets} shows how we generate a soft-clustering of datasets from historical query access logs using ICA. Algorithms~\ref{alg:assign_user_clusters} and \ref{alg:get_user_group_clusters} show how we associate users and groups of users with clusters. Algorithm~\ref{alg:get_candidate_tables} shows how we identify candidate tables for retrieval in a user's chat session.

\subsection{User Interface}
\label{ref:appendix_user_interface}
\begin{figure*}[h]
    \centering
    \includegraphics[width=\linewidth, keepaspectratio]{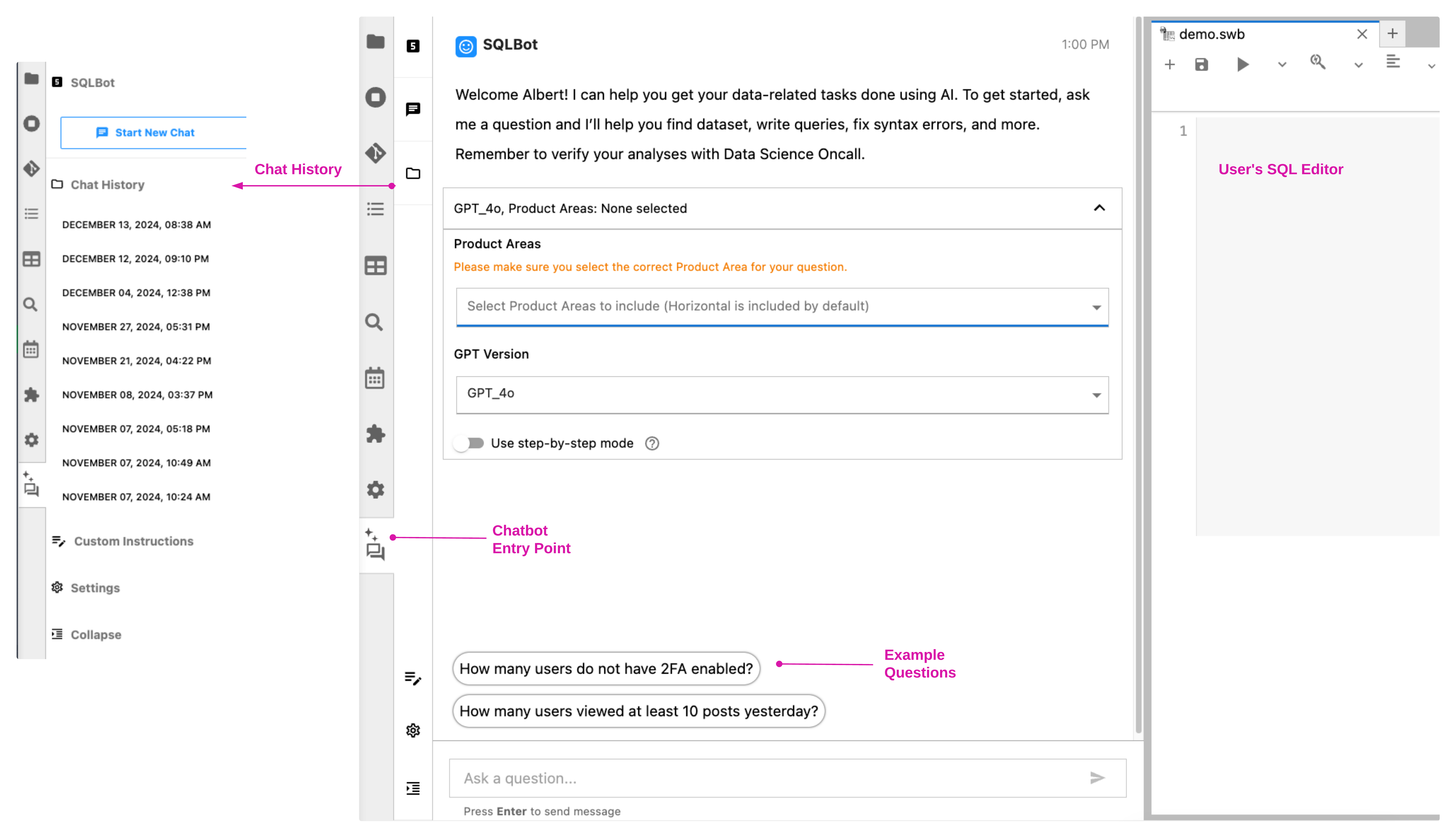}
    \caption{Chatbot is available in sidebar next to the SQL editor.}
    \label{fig:ui_chat}
    \Description[User interface showing the landing page of the chatbot.]{The entry point is in the sidebar with chat icon. There is a chat history button in the chatbot's menu, and example questions are shown right above the chatbox. The chatbot is contained within a sidebar of the SQL editor.}
\end{figure*}

Figure~\ref{fig:ui_chat} shows the chatbot's UI that the users can access through the company's internal platform for querying SQL. It also shows the different UI features like chat history, example questions, section to configure model and product areas.

\begin{figure*}[h]
    \vspace{-0.1in}
    \includegraphics[width=\linewidth, keepaspectratio]{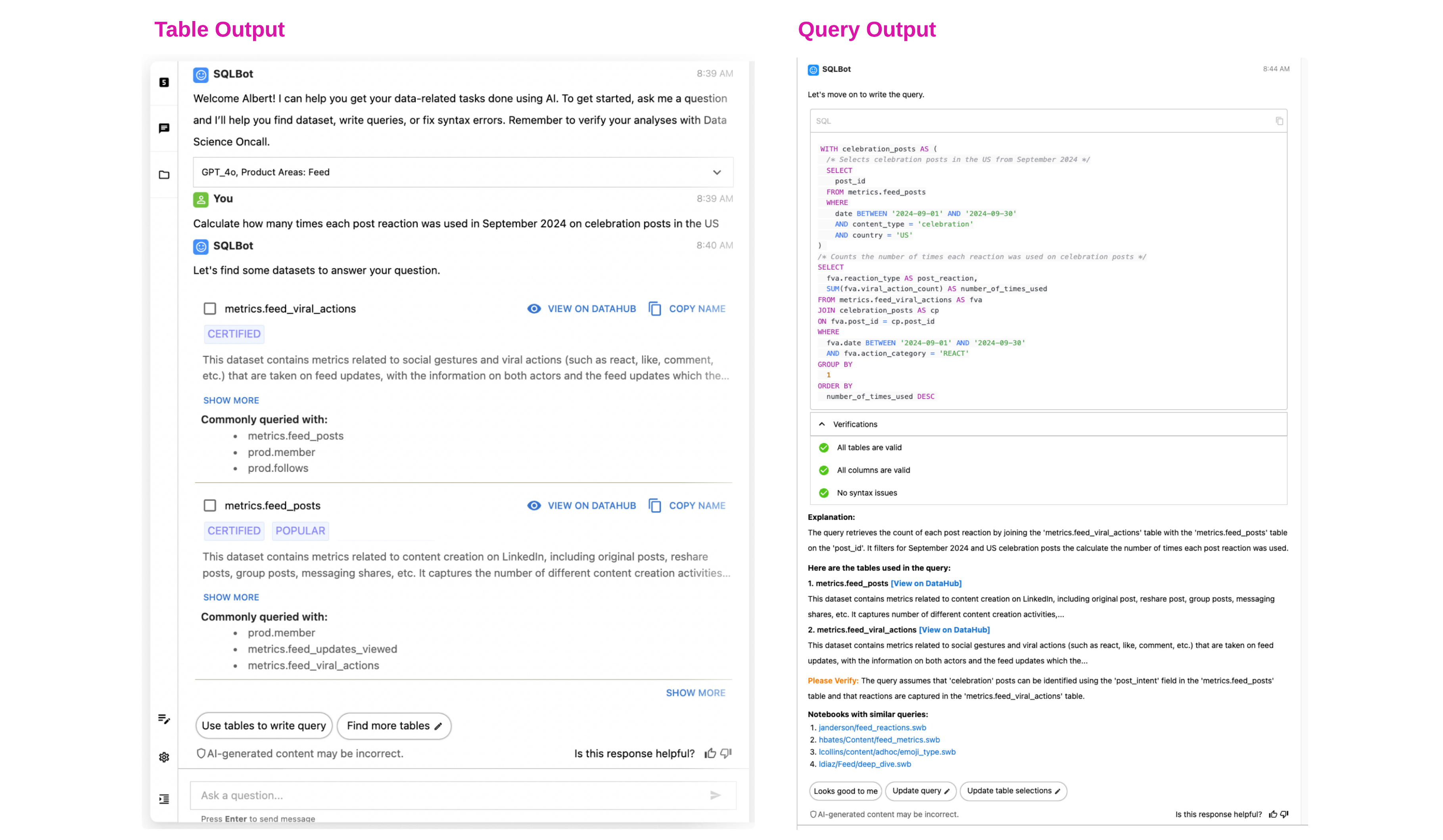}
    \caption{Rich display elements in query output help users understand responses and ask follow-up questions.}
    \label{fig:ui_output}
    \Description[Left: table output example. Right: query output example.]{The table output shows the chatbot's response to a request to find tables. It shows the table description, commonly queried tables, and tags such as ``certified'' and ``popular''. The query output shows the chatbot's response to a request to write a query. It shows the query with syntax highlighting and comments, verifications of query accuracy, explanation, tables used in the query, assumptions to verify, and links to notebooks with similar queries.}
    \vspace{-0.1in}
\end{figure*}

Figure~\ref{fig:ui_output} shows the Table Output and Query Output, each of which has rich set of information in addition to providing tables and SQL queries, to allow them to understand the response better.

\begin{figure*}[h]
    \centering
    \includegraphics[width=\linewidth]{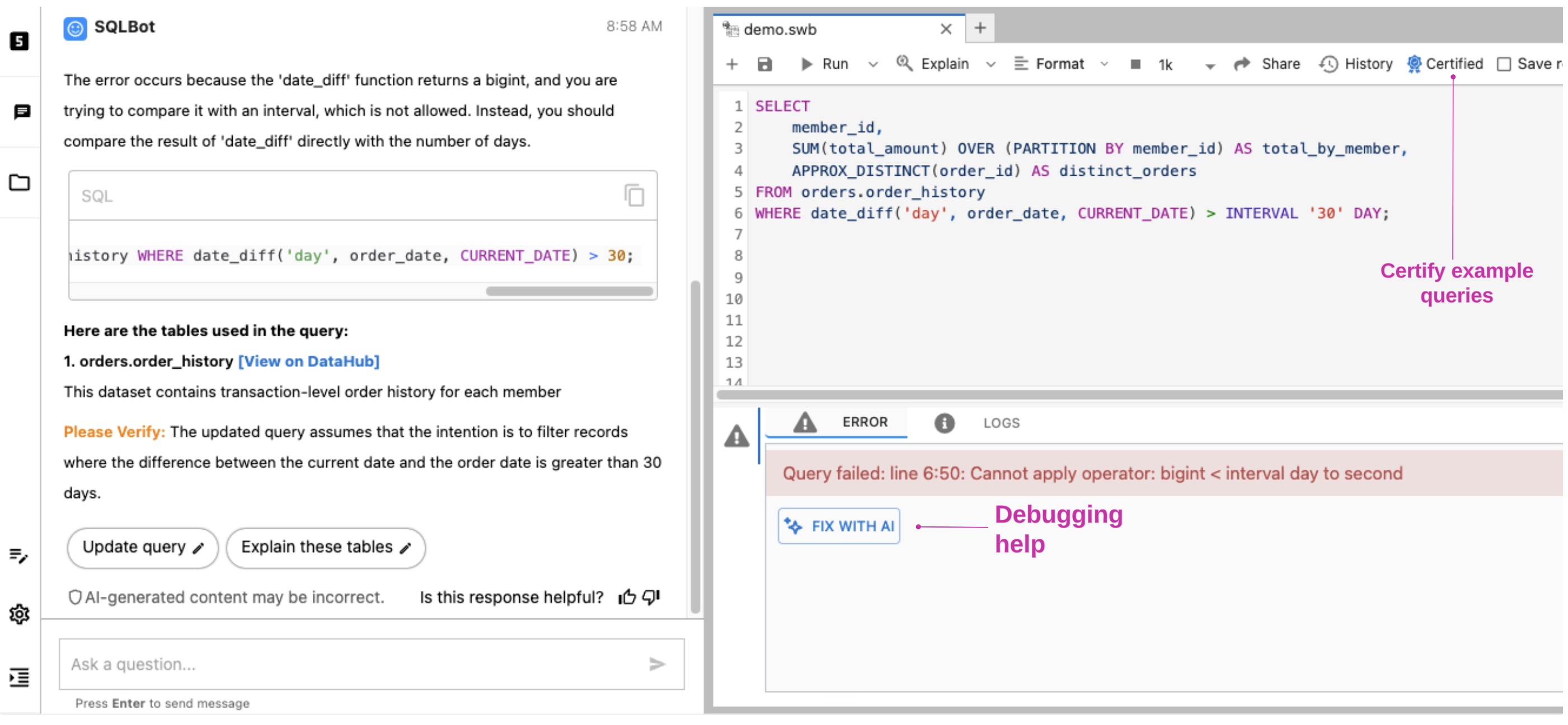}
    \caption{Fix-with-AI helps users debug query execution errors. Users can certify example queries from the query editor.}
    \label{fig:ui_fix}
    \Description[User interface showing a button to fix queries with AI.]{The button is shown right below the error message. The sidebar shows the chatbot's recommendation of how to fix the query. The top of the query editor includes a button to certify the queries in the current file.}
\end{figure*}

Figure~\ref{fig:ui_fix} shows the ``Fix with AI'' entry point to our chatbot, which appears whenever there is a query execution failure in the SQL editor. It also shows the button at the top of the SQL editor where users can mark a notebook as ``Certified'' to add example queries to the knowledge graph.

\subsection{Researcher Agent Architecture}
\label{ref:researcher_agent_architecture}
\begin{figure*}[h]
    \centering
    \includegraphics[width=0.8\linewidth]{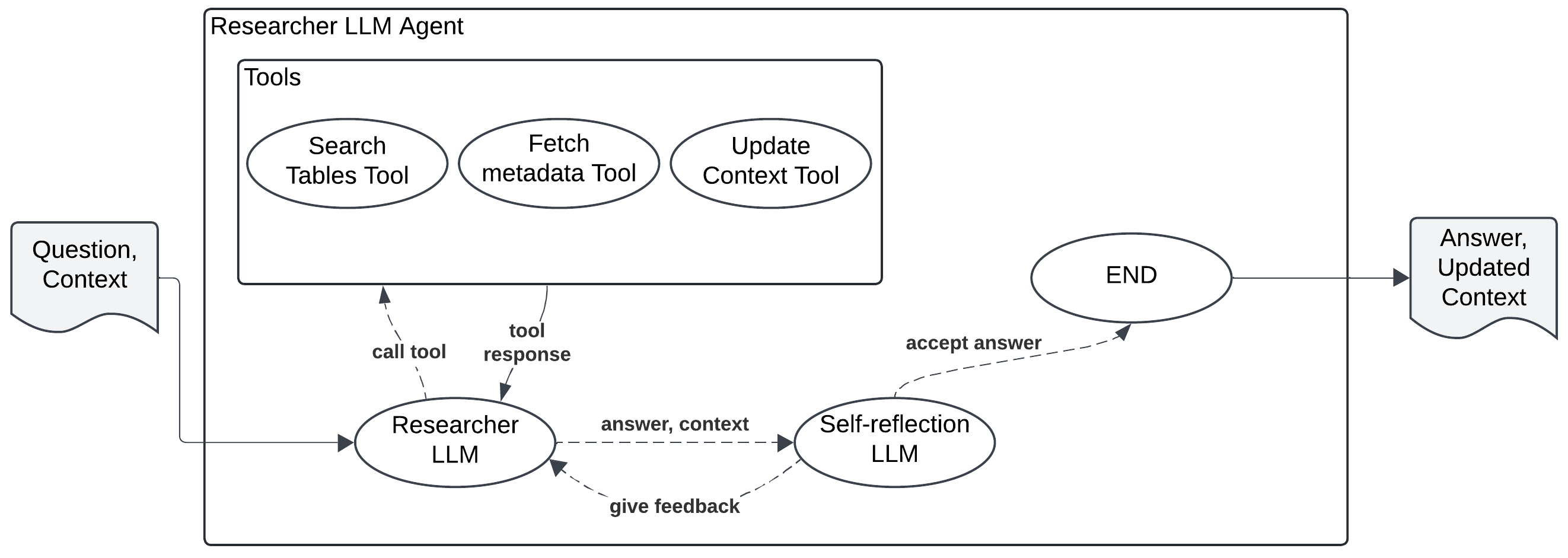}
    \caption{Researcher LLM Agent architecture. The Researcher is used within query fixing to search for tables to resolve hallucination.}
    \label{fig:researcher_llm_agent}
    \Description[The Researcher Agent uses tools to update the context.]{The question and current context are sent to Researcher LLM, which can call tools to search for tables, fetch their metadata, or update context. There is a self-reflection loop before returning the final answer.}
\end{figure*}

Figure~\ref{fig:researcher_llm_agent} shows how the Researcher LLM Agent calls tools and self-reflects to update tables in the context.

\subsection{Negative Results}
We experimented with a few ideas inspired by the literature that did not improve query accuracy in our setting.

Previous studies have reported the benefit of generating multiple queries and selecting the best one based on self-consistency of query outputs \cite{talaei_chess_2024, maamari_death_2024, lee2024mcs} or a selection model \cite{pourreza_chase-sql_2024}. In our setting, we do not have data access to run queries and need to return responses to users quickly. Therefore, we tried a related idea of introducing an extra LLM call before query writing that would generate up to three possibilities on how to write a query. These suggestions were provided to the query writer and can be interpreted as an explicit form of chain-of-thought prompting \cite{wei2022chain}. This did not improve recall on our benchmark. Our approach does not shuffle LLM inputs or incorporate results of query execution; while these are costly, they appear essential to the technique.

Following \citet{pourreza_din-sql_2023}, we introduced a query planner LLM call that decomposes the user question into more manageable tasks. The tasks are solved iteratively during query generation so that each new query builds upon the solutions to previous tasks. The query planner LLM returns the tasks in topological order with their parents so they can be solved sequentially. Adding these intermediate tasks incurs additional latency and we observed that they led to overly nested queries. Therefore we instructed the query planner to minimize the number of tasks and provided few-shot examples to explain when additional tasks are needed. The prompt for the query writer includes the plan from the planner along with solutions to any previous tasks. On our benchmark, the query planner lowered recall and quality metrics. Task breakdown constrains the solution generation process; we hypothesize it is not necessary for \texttt{gpt-4o}, although it might be useful for smaller models.

We also attempted to use the tasks from the planner to retrieve relevant context, as a form of query expansion. Tasks were used directly as query strings or concatenated with the user's question in the case of a single task. We only used tasks without a parent for retrieval, as those with parents typically use the same tables as their parents. However, this did not improve the recall of tables retrieved from EBR. A prompt that explicitly performs query expansion may prove more effective \cite{wang-etal-2023-query2doc, claveau2021neural, jagerman2023queryexpansionpromptinglarge}.

\begin{algorithm*}
\small
\caption{ClusterDatasets}\label{alg:cluster_datasets}
\begin{algorithmic}[1]
\Require matrix of user-dataset access counts
\Ensure Output dataset clusters and extended dataset clusters
\Function{ClusterDatasets}{$user\_dataset\_access$}

    \Comment{Applies ICA on users/datasets with sufficient total and unique accesses}
    \State $user\_dataset\_matrix \gets \Call{filter\_by\_usage}{user\_dataset\_access}$
    \State $scaled\_data \gets \Call{scale\_matrix}{user\_dataset\_matrix}$ \Comment{mean=0 and std=1 across users}
    \State $ica\_model \gets \Call{fit\_ica}{scaled\_data, n\_components=200}$ \Comment{we use \texttt{sklearn.decomposition.FastICA}}
    \State $ica\_output \gets ica\_model.\Call{transform}{scale\_data}$

    \Comment{Assigns datasets to clusters. Datasets can be in multiple clusters}
    \State $cluster\_dataset\_map$: map(str, list(str)) \Comment{maps cluster -> list(dataset)}
    \For{$cluster\_name$, $vector$ in $ica\_output$}
        \State $top\_datasets \gets vector\Call{.abs}{}()\Call{.sort}{DESC}\Call{.limit}{20}$ \Comment{20 datasets for each cluster}
        \State $cluster\_dataset\_map[cluster\_name] \gets top\_datasets$
    \EndFor

    \Comment{Assigns datasets to extended clusters. Every dataset is in at least one cluster}
    \State $cluster\_extended\_dataset\_map \gets \Call{copy}{cluster\_dataset\_map}$ \Comment{maps cluster -> list(dataset)}
    \For{$dataset\_name, vector$ in $\Call{transpose}{ica\_output}$}
        \State $closest\_cluster \gets vector\Call{.abs}{}()\Call{.sort}{DESC}\Call{.limit}{1}$
        \State $cluster\_extended\_dataset\_map[closest\_cluster]\Call{.append}{dataset\_name}$
    \EndFor
    \State \Return $cluster\_dataset\_map$, $cluster\_extended\_dataset\_map$
    \EndFunction
\end{algorithmic}
\end{algorithm*}

\begin{algorithm*}
\begin{minipage}{\textwidth}
\small
\caption{AssignUserClusters}\label{alg:assign_user_clusters}
\begin{algorithmic}[1]
\Require matrix of user-dataset access counts, dataset clusters
\Ensure Output top clusters for each user

    \Function{AssignUserClusters}{$user\_dataset\_access$, $cluster\_dataset\_map$}
    \State $user\_cluster\_access \gets user\_dataset\_access\Call{.map}{cluster\_dataset\_map}\Call{.groupby}{cluster}\Call{.sum}{}$ \Comment{User access counts for each cluster}
    \State $user\_cluster\_map$: map(str, list(str)) \Comment{maps user -> list(cluster)}
    \For{$user, cluster\_access$ in $user\_cluster\_access$}
        \State $cluster\_access \gets cluster\_access\Call{.where}{access \neq 0}$
        \State $user\_cluster\_map[user] \gets cluster\_access\Call{.sort}{DESC}\Call{.limit}{10}$
    \EndFor
    \State \Return $user\_cluster\_map$
    \EndFunction
\end{algorithmic}
\end{minipage}
\end{algorithm*}

\begin{algorithm*}
\begin{minipage}{\textwidth}
\small
\caption{GetUserGroupClusters}\label{alg:get_user_group_clusters}
\begin{algorithmic}[1]
\Require list of users, top clusters for each user
\Ensure Output list of top clusters for the users
    \Function{GetUserGroupClusters}{$users$, $user\_cluster\_map$}
        \State \Return $user\_cluster\_map\Call{.where}{user \in users}$
        \State \hspace{\algorithmicindent} $\Call{.groupby}{cluster\_name}$
        \State \hspace{\algorithmicindent} $\Call{.apply}{[\text{unique}, \text{sum}]}$
        \State \hspace{\algorithmicindent} $\Call{.sort\_values}{[\text{unique} \text{ DESC}, \text{sum} \text{ DESC}]}\Call{.limit}{K}$
    \EndFunction
\end{algorithmic}
\end{minipage}
\end{algorithm*}

\begin{algorithm*}
\begin{minipage}{\textwidth}
\small
\caption{GetCandidateTables}\label{alg:get_candidate_tables}
\begin{algorithmic}[1]
\Require user, list of product areas
\Ensure Output list of tables for the user and product areas
    \Function{GetCandidateTables}{$user$, $product\_areas$}
    \State $user\_clusters \gets \Call{GetUserGroupClusters}{user}$ \Comment{top clusters for the user}
    \State $product\_email\_groups \gets \Call{GetEmailGroups}{product\_areas}$ \Comment{email groups for the product areas}
    \State $representative\_users \gets \Call{GetEmployees}{email\_groups}$ \Comment{employees in those groups}
    \State $product\_area\_clusters \gets \Call{GetUserGroupClusters}{representative\_users}$
    \State $clusters \gets \Call{merge}{user\_clusters, product\_area\_clusters}$
    \State $inferred\_tables \gets \Call{GetExtendedTables}{clusters}$ \Comment{extended tables from the clusters}
    \State $explicit\_tables \gets \Call{GetExplicitTables}{product\_areas}$ \Comment{tables explicitly added to the product areas}
    \State \Return $\Call{merge}{inferred\_tables, explicit\_tables}$
    \EndFunction
\end{algorithmic}
\end{minipage}
\end{algorithm*}

\end{document}